\documentclass{article}
\usepackage{xcolor}
\PassOptionsToPackage{hyperfootnotes=false}{hyperref}
\usepackage[preprint]{corl_2026}
\usepackage[T1]{fontenc}
\usepackage[bottom]{footmisc}
\usepackage{amsmath,amssymb,booktabs,tabularx,array,graphicx}
\hypersetup{pdftitle={Evaluation-Guided Demonstration Refinement for Vision-Language-Action Policies: The Winning Entry in the MetaFine Open Competition},pdfauthor={Lifan Zheng, Guotao Chen, Zhenyang Li, Shujie Cui, Qiang Huang, Fan Liu}}
\newcommand{\pifive}{$\pi_{0.5}$}
\newcolumntype{Y}{>{\raggedright\arraybackslash}X}
\title{Evaluation-Guided Demonstration Refinement\\for Vision-Language-Action Policies\\[4pt]{\normalfont\large The Winning Entry in the MetaFine Open Competition}}
\author{\textbf{Lifan Zheng$^{1}$\quad Guotao Chen$^{1}$\quad Zhenyang Li$^{2}$\quad Shujie Cui$^{3}$\quad Qiang Huang$^{4}$\quad Fan Liu$^{1,}$}\thanks{Corresponding author.}\\[4pt]
{\normalfont $^{1}$Southeast University\quad $^{2}$The Hong Kong Polytechnic University}\\
{\normalfont $^{3}$Monash University\quad $^{4}$Harbin Institute of Technology (Shenzhen)}\\[3pt]
{\normalfont\small\texttt{\{z1ivan,\,liufan\}@seu.edu.cn}}\\[3pt]
{\normalfont\small Technical report for RoboFineMani @ CoRL 2026}}
\begin{document}
\maketitle

\begin{abstract}
We fine-tuned a separate \pifive{} policy for each of the five tasks in the MetaFine Open Competition, with a task-specific training mixture and number of actions executed per prediction. Training started from the official demonstrations and expert procedures. Local evaluation exposed recurring grasping, pushing, and insertion failures. We addressed them by revising the demonstrated motions and by adding expert continuations from states the policy reached, while keeping nominal demonstrations in training. Checkpoints were compared on every instruction, so that a gain on one instruction could not hide a loss on another. Our submission, MAI, ranked first overall with a score of 96.66.
\end{abstract}

\section{Introduction}
MetaFine~\citep{xu2026beyond} evaluates fine-grained manipulation through instruction understanding, perceptual robustness, and behavior. The MetaFine Open Competition~\citep{metafinechallenge} covers five tabletop tasks: bottle-part grasping (T1), mug grasp-and-move (T2), switch toggling (T3), block sorting (T4), and letter insertion (T5). Understanding $U$ measures success across instructions; perception $P$ averages normalized areas under camera and lighting degradation curves; behavior $B$ measures task success, or normalized stage completion for the multi-stage T2 and T4. Each task receives a composite score $S=(P+U+B)/3$.

We fine-tuned one \pifive{}~\citep{intelligence2025pi_} policy for each task. We used MetaFine's per-instruction scores, together with our own grouping of episodes by scene layout, to choose which trajectories to inspect, and revised the demonstrations where those trajectories showed recurring errors. The submitted policies scored 96.66 overall and ranked first on the final leaderboard.

\section{Technical Approach}
\label{sec:approach}
\subsection{Policy and training setup}
We fine-tuned all parameters of \pifive{} using OpenPI~\citep{openpi} and the flow-matching behavior-cloning objective~\citep{black2024pi_0,Lipman_2022_221002747}. The inputs are the instruction, seven arm and two finger joint positions, and RGB images from the base and wrist cameras. Each predicted action contains seven joint deltas and a gripper command. The policy generates a chunk of 50 actions using ten denoising steps, then executes the first 20 actions for T1, 30 for T3, or 10 for T2/T4/T5 before replanning.

We trained with AdamW~\citep{Loshchilov_2017_171105101}, learning-rate warmup, and cosine decay. In the final training rounds, T2 started from the base model and each other task from its previous policy. Checkpoints were selected by local evaluation (Section~\ref{sec:results}).

\subsection{Refining the official demonstrations}
Training used the official demonstrations and trajectories generated by the official expert procedures. We ran these procedures on additional scenes and revised them in response to the failures found during development. Only successful demonstrations with consistent recorded observations and actions entered training. Where demonstrations contained unnecessary pauses or repeated approaches, as in T1 and T5, we shortened or replanned them. Variation in object pose and task geometry was kept, as were the motions needed to finish the task.

Some failed rollouts reached states that nominal demonstrations rarely cover. From such states the expert recorded continuations, following the rationale of dataset aggregation~\citep{ross2011reduction}. We also built recovery demonstrations from constructed failure configurations, such as a grasp on the wrong bottle part in T1, and excluded the actions used to set up these configurations from training. Expert procedures could use simulator state; deployed policies received only images, joint positions, and language. When continuations or recovery demonstrations were used, task-specific source quotas kept nominal demonstrations in every training batch.

\subsection{Training for each task}
Table~\ref{tab:tasks} summarizes the observed behavior, training approach, and evaluation checks for each task.

\begin{table}[!ht]
\caption{Observed behavior, training approach, and evaluation checks for each task.}
\label{tab:tasks}
\centering\small
\begin{tabularx}{\linewidth}{@{}p{7mm}YYY@{}}
\toprule
Task & Observed behavior & Training approach & Evaluation checks \\
\midrule
T1 & Part-dependent failures; slips as lifting begins & Part-specific grasps; gradual lift during closure & Cap and body scored separately \\
T2 & All nominal cases succeed; two perturbation failures & Balanced sources and motion directions & All directions; visual perturbations \\
T3 & Missed contact and partial pushes at difficult angles & Complete pushes; recovery across approach angles & Both switch colors; reserve scene set \\
T4 & Empty grasps near close neighbors & Neighbor clearance and empty-grasp recovery & All colors; standard and crowded scenes \\
T5 & Letter o released offset or unseated & Continuous plans with consistent grasps; full seating & All four letters; two scene sets \\
\bottomrule
\end{tabularx}
\end{table}

\paragraph{T1: grasping the instructed bottle part.}
Local rollouts showed slips as lifting began. We adjusted the trajectories for both the cap and body instructions, shortening hesitation and beginning a gradual lift during closure. Nominal demonstrations, expert continuations, and recovery demonstrations remained in training. We evaluated the two instructions separately because earlier variants had improved body success at the expense of cap success.

\paragraph{T2: mug grasp-and-move.}
Batches balanced official demonstrations with trajectories generated by the official expert on additional scenes, with equal coverage of the three motion directions. We kept this recipe for the final submission after checking nominal and perturbed conditions.

\paragraph{T3: switch toggling.}
Local failures included missed contact and incomplete toggles at difficult approach angles. We revised the approach and extended the push through the toggle, with recovery demonstrations for failed contact. Later training added approaches that pass above the switch center, alongside the earlier demonstration sources.

\paragraph{T4: block sorting.}
In crowded layouts, the gripper sometimes closed without the target block. We adjusted the expert approach to leave clearance from neighboring blocks and recorded expert continuations from these empty-grasp states. Training mixed these trajectories with the official and earlier expert demonstrations.

\paragraph{T5: letter insertion.}
Earlier rounds replaced the T5 demonstrations with continuous planned trajectories that use one consistent grasp per letter. The remaining failures on o involved lateral offsets and release before seating. We revised the o trajectories to align the letter above the opening and seat it fully before release. The revised o demonstrations included trajectories with varied in-hand offsets, expert continuations from policy rollouts, and recoveries from a letter resting on the board. They replaced the earlier o demonstrations; C/R/L demonstrations were unchanged.

\section{Competition Results}
\label{sec:results}
\subsection{Official evaluation}
Our submission scored 96.66 overall and ranked first on the organizers' final leaderboard; the second- and third-place scores were 94.71 and 91.37~\citep{metafineleaderboard}. Table~\ref{tab:official} lists the scores for each task. T1 had the lowest score, and overall perception ($P=95.47$) was lower than understanding ($U=97.25$).

\begin{table}[!ht]
\caption{Official results (\%). Overall averages the five tasks.}
\label{tab:official}
\centering\small
\begin{tabular}{@{}lcccccc@{}}
\toprule
 & T1 & T2 & T3 & T4 & T5 & Overall \\
\midrule
$P$ & 86.25 & 98.89 & 98.54 & 99.31 & 94.37 & 95.47 \\
$U$ & 90.00 & 100.00 & 100.00 & 100.00 & 96.25 & 97.25 \\
$B$ & 90.00 & 100.00 & 100.00 & 100.00 & 96.25 & 97.25 \\
\midrule
$S$ & 88.75 & 99.63 & 99.51 & 99.77 & 95.62 & 96.66 \\
\bottomrule
\end{tabular}
\end{table}

\subsection{Local development comparisons}
Local evaluation used the competition's task predicates, action interface, and scoring~\citep{metafinechallenge}. Table~\ref{tab:local} reports local scores for the submitted policies, with reference comparisons for T1 and T3--T5. Each task used its own evaluation suite; we do not average the local scores across tasks.

\begin{table}[!ht]
\caption{Local composite scores $S$ (\%): Selected denotes the final submitted policy; Reference denotes a preceding task-specific policy evaluated on the same scene set. T2 was trained from the base model; ``---'' means no reference comparison is reported. $n$ counts episodes per model. For T2, $S$ follows the public competition configuration, which evaluates $P$ on the left instruction; the sweep also covers right and forward.}
\label{tab:local}
\centering\small
\begin{tabularx}{\linewidth}{@{}lYrr@{}}
\toprule
Task & Evaluation scope & Reference & Selected \\
\midrule
T1 & 80 scenes; $n=3{,}200$ & 88.95 & 93.39 \\
T2 & 20 scenes; $n=480$ & --- & 100.00 \\
T3 & 120 scenes, four inference seeds; $n=7{,}680$ & 99.22 & 99.47 \\
T4 & 80 scenes; $n=1{,}280$ & 99.81 & 99.97 \\
T5 & 80 development scenes; $n=880$ & 95.07 & 97.95 \\
 & Separate 80 scenes; $n=880$ & 93.26 & 97.05 \\
\bottomrule
\end{tabularx}
\end{table}

T1 improved on both instructions, with a larger gain in cap success. On 80 scenes, $S$ rose from 88.95 to 93.39. Cap success $U$ increased from 88.96 to 97.71, while body success rose from 88.96 to 90.21 and remained the weaker instruction. The scene-clustered 95\% interval for the difference in $S$ was $[1.24,8.11]$ points.

T2 succeeded in all 60 nominal instruction cases; the 480-episode sweep contained two failures, both under visual perturbations of the forward instruction. Under the standard scoring scope (Table~\ref{tab:local}), $S$ was 100.00; extending $P$ to all three directions gave $S=99.81$.

The earlier T3 revisions produced large gains, but the final revision did not establish a further improvement. Across two rounds, $S$ on the 80-scene development suite rose from 74.9 to 91.6 and then 99.2. That policy is the reference in Table~\ref{tab:local}. We compared it with the selected policy on 120 scenes: the development suite, a 20-scene reserve set used to compare checkpoints, and 20 further scenes. With four inference seeds per model, the selected policy scored just 0.24 points higher (computed before rounding). The 95\% scene-bootstrap interval, $[-0.26,0.79]$, includes zero, leaving the difference uncertain.

T4's clearest gain was in crowded layouts: successes increased from 313/348 to 348/348 on 116 scenes, eliminating the 35 failures observed with the reference. Both policies already had near-ceiling standard-suite $S$ (Table~\ref{tab:local}). The selected policy also succeeded in all 720 nominal instruction episodes of the standard 80-scene suite and failed only once in 4,452 episodes on a separate 1,484-scene set. The larger-set result is reported without a reference comparison.

T5 improved chiefly on o, the letter targeted by the final revisions. Success on o rose from 82.5 to 97.5 on the development set and from 87.5 to 97.5 on the separate set; $S$ also increased on both. The overall gains came with a small regression on C: development-set success fell from 100 to 98.75, despite its demonstrations being unchanged.

\section{Lessons Learned and Discussion}
\label{sec:lessons}
\textbf{Per-instruction scores and scene grouping revealed failures hidden by the task score.} MetaFine's separate instruction scores distinguished cap from body in T1 and isolated the letter o in T5. In T4 the task score was already high, and our own grouping of episodes by layout showed that the remaining failures concentrated near neighboring blocks.

\textbf{Inspecting the few steps around contact helped distinguish failures that scores alone could not explain.} In T1, T3, and T5, we observed a grasp that closed and then slipped, a push that stopped short, and a letter released above the slot, respectively. In response, we revised those segments of the demonstrations: a gradual lift during closure for T1, a longer push for T3, and full seating before release for T5.

\textbf{Targeted revisions could introduce regressions on other instructions or scenes.} A T3 variant trained without the earlier sources regressed on the reserve set and was rejected, so the final mixture kept them. Keeping an instruction's demonstrations did not always preserve its performance: earlier T1 variants improved body grasping but lost cap success, and the selected T5 policy declined slightly on C. We therefore checked individual instructions when selecting checkpoints, even when the task score improved.

\textbf{Repeated model selection and combined training changes limit what the local comparisons establish.} The local sets served both to guide revisions and to select checkpoints, so their scores describe development performance and the reported intervals do not account for that selection. Official T1 and T5 scores were below their local values, while T2--T4 were within half a point; because the official and local suites use different scenes, these gaps do not identify a cause. Training rounds also combined demonstration and sampling changes, so the local comparisons do not isolate the effect of either.

\paragraph{Acknowledgments.}
We thank the MetaFine authors and organizers for the benchmark and competition resources, and the OpenPI contributors for their code.

\bibliography{references}
\end{document}